\documentclass[10pt,twocolumn,letterpaper]{article}

\usepackage{iccv}
\usepackage{times}
\usepackage{epsfig}
\usepackage{graphicx}
\usepackage{amsmath}
\usepackage{amssymb}

\usepackage{caption}
\usepackage{subcaption}

\usepackage{multicol}
\usepackage{multirow}

\usepackage{algorithm,algorithmic}
\usepackage{cuted}

\usepackage{capt-of}
\usepackage{overpic}

\newcommand{\eR}{\mathbb{R}}

\newcommand{\eeAT}{A^{\dag}}

\newcommand{\iee}{\emph{i.e.}}
\newcommand{\egg}{\emph{e.g.}}
\newcommand{\ieB}{\emph{i.e. }}
\newcommand{\egB}{\emph{e.g. }}

\newcommand{\loss}{\mathcal{L}}

\newcommand{\range}{\mathcal{R}_A}
\newcommand{\nullA}{\mathcal{N}_A}

\newcommand{\ntransf}{|\mathcal{G}|}
\newcommand{\group}{\mathcal{G}}

\usepackage[breaklinks=true,bookmarks=false]{hyperref}

\iccvfinalcopy 

\def\iccvPaperID2251Enter the ICCV Paper ID here
\def\httilde{\mbox{\tt\raisebox{-.5ex}{\symbol{126}}}}

\begin{document}
\title{Equivariant Imaging: Learning Beyond the Range Space}

\author{Dongdong Chen\\
School of Engineering\\
University of Edinburgh\\
\and
Julián Tachella \\
School of Engineering\\
University of Edinburgh\\
\and
Mike E. Davies\\
School of Engineering\\
University of Edinburgh\\
}

\maketitle


\begin{strip}
\vskip -6em
\centering
\includegraphics[width=0.99\textwidth]{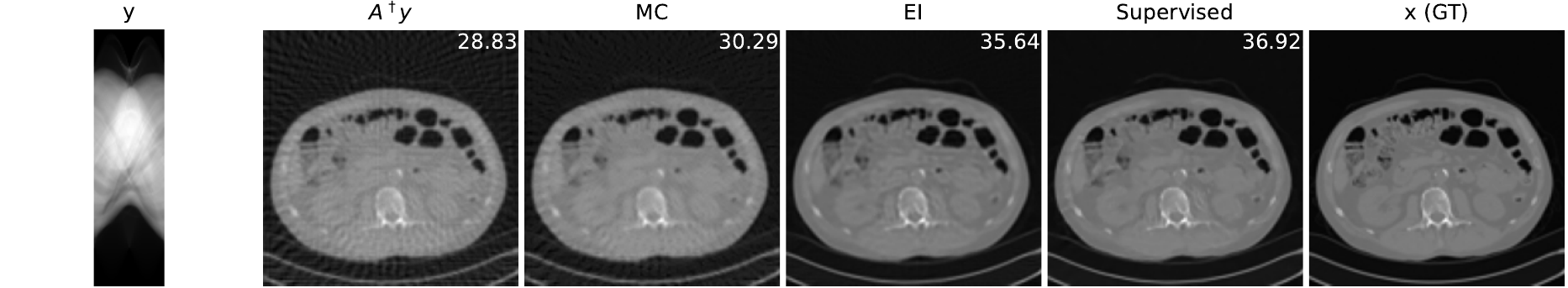}
\includegraphics[width=0.99\textwidth]{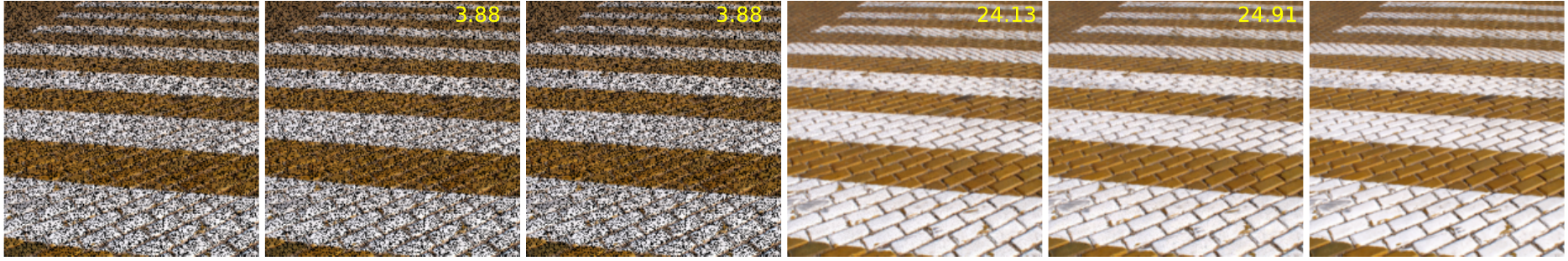}
\captionof{figure}{\textbf{Learning to image from only measurements}. Training an imaging network through just measurement consistency (MC) does not significantly improve the reconstruction over the simple pseudo-inverse ($A^\dagger y$). However, by enforcing invariance in the reconstructed image set, \emph{equivariant imaging} (EI) performs almost as well as a fully supervised network. \textbf{Top}: sparse view CT reconstruction, \textbf{Bottom}: pixel inpainting. PSNR is shown in top right corner of the images.} \label{fig:cover results}
\end{strip}

\begin{abstract}
In various imaging problems, we only have access to compressed measurements of the underlying signals, hindering most learning-based strategies which usually require pairs of signals and associated measurements for training. Learning only from compressed measurements is impossible in general, as the compressed observations do not contain information outside the range of the forward sensing operator.
We propose a new end-to-end self-supervised framework that overcomes this limitation by exploiting the equivariances present in natural signals. Our proposed learning strategy performs as well as fully supervised methods. Experiments demonstrate the potential of this framework  on inverse problems including sparse-view X-ray computed tomography on real clinical data and image inpainting on natural images. Code has been made available at: \url{https://github.com/edongdongchen/EI}.
\end{abstract}

\vspace{-6pt}

\section{Introduction}
Linear inverse problems are ubiquitous in computer vision and signal processing, appearing in multiple important applications such as super-resolution, image inpainting and  computed tomography (CT).
The goal in these problems consists of recovering a signal $x$ from measurements $y$, that is inverting the forward process
\begin{equation}\label{eqs:base}
  y = Ax + \epsilon,
\end{equation}
which is generally a challenging task due to the ill-conditioned operator $A$ and noise $\epsilon$.
In order to obtain a stable inversion, traditional approaches have used linear reconstruction, \ieB $A^\dagger y$, where the estimate is restricted to the range space of $A^{\top}$, or model-based approaches that reduce the set of plausible reconstructions $x$ using prior information (\egB sparsity).
Leveraging the powerful representation properties of deep neural networks, a different approach is taken by recent end-to-end learning solutions which learn the inverse mapping directly from samples $(x,y)$.
However, all of these approaches require ground truth signals $x$ for learning the reconstruction function $x=f(y)$, which hinders their applicability in many real-world scenarios where ground truth signals are either impossible or expensive to obtain. This limitation raises the natural question: \emph{can we learn the reconstruction function without imposing strong priors and without knowing the ground-truth signals}?

Here, we show that typical properties of physical models such as rotation or shift invariance, constitute mild prior information that can be exploited to learn beyond the range space of $A^{\top}$. We present an end-to-end \emph{equivariant imaging} framework which can learn the reconstruction function from compressed measurements $y$ alone for a single forward operator.
As shown in Figure \ref{fig:cover results}, the equivariant imaging approach performs almost as well as having a dataset of ground truth signals $x$ and significantly outperforms simply enforcing the measurement consistency $Af(y)=y$ in the training process.
We show both theoretically and empirically that the proposed framework is able to identify the signal set and learn the reconstruction function from few training samples of compressed observations without accessing ground truth signals $x$.  Experimental results demonstrate the potential of our framework through qualitative and quantitative evaluation on inverse problems. Specifically our contributions are as follows:

\begin{enumerate}
    \item We present a conceptually simple \emph{equivariant imaging} paradigm for solving inverse problems without ground truth. We show how invariances enable learning beyond the range space of the adjoint of the forward operator, providing necessary conditions to learn the reconstruction function.
    \item We show that this framework can be easily incorporated into deep learning pipelines using an additional loss term enforcing the system equivariance.
    \item We validate our approach on sparse-view CT reconstruction and image inpainting tasks, and show that our approach obtains reconstructions comparable to fully supervised networks trained with ground truth signals.
\end{enumerate}

\subsection{Related work}

\paragraph{Model-based approaches}
The classical model-based approach for solving inverse problems \cite{boyd2011distributed,daubechies2004iterative}, constrains the space of plausible solutions using a fixed model based on prior knowledge (\egg~sparsity \cite{ng2010solving}). 
Although the model-based paradigm  has typically nice theoretical properties, it presents two disadvantages: constructing a good prior model that captures the low-dimensionality of natural signals is generally a challenging task. Moreover, the reconstruction can be computationally expensive, since it requires running an optimization procedure at test time.

\paragraph{Deep learning approaches}
Departing from model-based strategies, the deep learning strategies aim to learn the reconstruction mapping from samples $(x,y)$. This idea has been successfully applied to a wide variety of inverse problems, such as image denoising and inpainting \cite{mao2016image,yu2019free,yu2018generative}, super-resolution \cite{dong2015image,wang2018esrgan,lugmayr2020srflow}, MRI reconstruction \cite{mardani2018neural,chen2020compressive} and CT image reconstruction \cite{jin2017deep,yang2018low}. However, all of these approaches require access to training pairs $(x,y)$ which might not be available in multiple real-world scenarios.

\paragraph{Learning with compressed observations}
In general, given a fixed forward model $A$ with a non trivial nullspace, it is fundamentally impossible to learn the signal model beyond the range space of $A^{\top}$ using only compressed samples $y$. This idea traces back to \textit{blind compressive sensing}~\cite{gleichman2011blind}, where it was shown that is impossible to learn a dictionary from compressed data without imposing strong assumptions on the set of plausible dictionaries.

\paragraph{Self-supervised learning}
More recently, there is a growing body of work on \emph{self-supervised} learning exploring what can be learnt without ground truth. For example, there is a collection of studies in the mould of Noise2X \cite{lehtinen2018noise2noise,batson2019noise2self,krull2019noise2void,moran2020noisier2noise} where image denoising is performed without access to the ground truth. However, the denoising task does not have a non trivial nullspace since $A$ is the identity. Although some follow-up works including \cite{hendriksen2020noise2inverse} and \cite{liu2020rare} have tried to solve a more general situation, the former does not consider a nontrivial null space while the latter requires the exploitation of the diversity of multiple forward operators to learn a denoiser and eventually solves the inverse problem in an iterative model-based optimization \cite{romano2017little}. Finally, some unconditional~\cite{bora2018ambientgan} and conditional~\cite{pajot2019unsupervised} generative models, were proposed to learn to reconstruct from compressed samples, but again requiring multiple different forward operators. In contrast, we are able to learn this for a single forward operator.

\section{Method}

\subsection{Problem Overview}
We consider a linear imaging physics model $A: \mathbb{R}^n\rightarrow \mathbb{R}^m$, and the challenging setting in which only a set of $N$ compressed observations $\{y_i\}_{i=1,\dots,N}$ are available for training. The learning task consists of learning a reconstruction function $f_\theta:\mathbb{R}^m\rightarrow \mathbb{R}^n$ such that $f_\theta(y)=x$. As the number of measurements is lower than the dimension of the signal space $m<n$, and the operator $A$ has a non trivial nullspace.

\paragraph{Measurement consistency}
Given that we only have access to compressed data $y$, we can enforce that the inverse mapping $f$ is consistent in the measurement domain. That is
\begin{equation}
\label{eq:data consistency}
    Af(y) = y.
\end{equation}
However, this constraint is not enough to learn the inverse mapping, as it cannot capture information about $\mathcal{X}$ outside the range of the operator $A^{\top}$. As shown in Section~\ref{sec:theory}, there are multiple functions $f_\theta$ that can verify \eqref{eq:data consistency}, even if we have infinitely many samples $y_i$.

\paragraph{Invariant set consistency}
In order to learn beyond the range space of $A^{\top}$, we can exploit some mild prior information about the set of plausible signals $\mathcal{X}$. We assume that the set presents some symmetries, \ieB that it is invariant to certain groups of transformations, such as shifts, rotations, reflections, etc. This assumption has been widely adopted both in multiple signal processing and computer vision applications. 
For example, it is commonly assumed that natural images are shift invariant. Another example is computed tomography data, where the same organ can be imaged at different angles making the problem invariant to rotation.


Under this assumption, the set of signals $\mathcal{X}$ is invariant to a certain group of transformations $\group =\{ g_1,\dots,g_{\ntransf} \}$ which are unitary matrices
$T_g \in \eR^{n\times n}$  such that for all $x\in \mathcal{X}$, we have
\begin{equation}\label{eq:ti}
    T_g x \in \mathcal{X}
\end{equation}
 for $\forall g\in \group$, and the sets $T_g\mathcal{X}$ and $\mathcal{X}$ are the same.

According to the invariance assumption in \eqref{eq:ti},  the composition $f\circ A$ should then be equivariant to the transformation $T_g$, \iee
\begin{equation}\label{eq:equivariant set consistency}
    f(AT_gx) = T_gf(Ax)
\end{equation}
for all $x\in \mathcal{X}$, and all $g\in \group$. It is important to note that \eqref{eq:equivariant set consistency} does not require $f$ to be invariant or equivariant, but rather the composition $f\circ A$ to be equivariant.
As discussed in Section~\ref{sec:theory}, as long as the range of $A^{\top}$ itself is not invariant to all $T_g$, this additional constraint on the inverse mapping $f$ allows us to learn beyond the range space.

\paragraph{Invariant distribution consistency}
In most cases, not only is the signal set $\mathcal{X}$ invariant but also the  distribution $p(x)$ defined on this set is invariant, i.e.
\begin{equation}
\label{eq:distrib_inv}
    p(T_gx) = p(x)
\end{equation}
for all $g\in\group$. Hence, we can also enforce this distributional constraint when learning the inverse mapping $f$.

\begin{algorithm}[t]
\algsetup{linenosize=\tiny}
\scriptsize
\ttfamily\textcolor{teal}{}\\
\ttfamily\textcolor{teal}{\#\hspace{1.5mm}A.forw, A.pinv:\hspace{1.5mm}forward and pseudo inverse operators}\\
\ttfamily\textcolor{teal}{\#\hspace{1.5mm}G:\hspace{1.5mm}neural network}\\
\ttfamily\textcolor{teal}{\#\hspace{1.5mm}T:\hspace{1.5mm}transformations group}\\
\ttfamily\textcolor{teal}{\#\hspace{1.5mm}a:\hspace{1.5mm}alpha}\\

\ttfamily{for y in loader:}\ttfamily\textcolor{teal}{\hspace{1.5mm}\#\hspace{1.5mm}load a minibatch y with N samples}\\
\hspace*{4.3mm}\ttfamily\textcolor{teal}{\#\hspace{1.5mm}randomly select a transformation from T}\\
\hspace*{4.3mm}\ttfamily{t = select(T)}\\
\\
\hspace*{4.3mm}\ttfamily{x1 = G(A.pinv(y))}\ttfamily\textcolor{teal}{\hspace{1.5mm}\#\hspace{1.5mm}reconstruct x from y}\\
\hspace*{4.3mm}\ttfamily{x2 = t(x1)}\ttfamily\textcolor{teal}{\hspace{1.5mm}\#\hspace{1.5mm}transform x1}\\
\hspace*{4.3mm}\ttfamily{x3 = G(A.pinv(A.forw(x2)))}\ttfamily\textcolor{teal}{\hspace{1.5mm}\#\hspace{1.5mm}reconstruct x2}\\
\\
\hspace*{4.3mm}\ttfamily\textcolor{teal}{\#\hspace{1.5mm}training loss, Eqn.(\ref{eqs:ten_loss})}\\
\hspace*{4.3mm}\ttfamily{loss = MSELoss(A.forw(x1), y)}\ttfamily\textcolor{teal}{\hspace{1.5mm}\#\hspace{1.5mm}data consistency}\\
\hspace*{11.75mm}\ttfamily{+ alpha*MSELoss(x3, x2)}\ttfamily\textcolor{teal}{\hspace{1.5mm}\#\hspace{1.5mm}equivariance}\\
\\
\hspace*{4.3mm}\ttfamily\textcolor{teal}{\#\hspace{1.5mm}update G network}\\
\hspace*{4.3mm}\ttfamily{loss.backward()}\\
\hspace*{4.3mm}\ttfamily{update(G.params)}\\
\caption{Pseudocode of EI in a PyTorch-like style.}\label{algo:ten}
\end{algorithm}

\subsection{Equivariant Imaging}
We propose to use a trainable deep  neural network $G_\theta: \eR^n\rightarrow \eR^n$ and an approximate linear inverse (for example a pseudo-inverse) $A^{\dagger}\in \eR^{n\times m}$ to define the inverse mapping as $f_{\theta} = G_{\theta}\circ A^{\dagger}$.
We emphasize that while in principle the form of $f_\theta$ is flexible, here we use the linear $A^\dagger$ to first project $y$ into $\mathbb{R}^n$ to simplify the learning complexity. In practice $A^\dagger$ can be chosen to be any approximate inverse that is cheap to compute.

\begin{figure}[h]
\begin{center}
\includegraphics[width=1\linewidth]{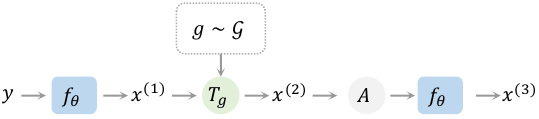}
\end{center}
\caption{\textbf{Equivariant learning strategy.} $x^{(1)}$ represents the estimated image, while $x^{(2)}$ and $x^{(3)}$  represent $T_gx^{(1)}$ and the estimate of $x^{(2)}$ from the measurements $\tilde{y} = A x^{(2)}$ respectively.}
\label{fig:forward pass}
\end{figure}

We propose a training strategy that enforces both the measurement consistency in~\eqref{eq:data consistency} and the equivariance condition in~\eqref{eq:equivariant set consistency} using only a dataset of compressed samples $\{y_i\}_{i=1,\dots,N}$.
In the forward pass, we first compute $x^{(1)}=f_\theta(y)$ as an estimate of the actual ground truth $x$ which is not available for learning. Note the data consistency between $Af_\theta(y)$ and $y$ only ensures that $Ax^{(1)}$ stays close to the input measurement $y$ but fails to learn beyond the range space of $A^{\top}$. According to the equivariance property  in \eqref{eq:equivariant set consistency}, we subsequently transform $x^{(2)}=T_g x^{(1)}$, for some $g\sim \group$, and pass it through $f\circ A$ to obtain $x^{(3)}$.
 The computations of $x^{(1)}$, $x^{(2)}$ and $x^{(3)}$ are illustrated in Figure~\ref{fig:forward pass}.

The network weights are updated according to the error between $y$ and $Ax^{(1)}$, and the error between $x^{(2)}$ and $x^{(3)}$,
by minimizing the following training loss
\begin{equation}\label{eqs:ten_loss}
      \arg\min_\theta \mathbb{E}_{y,g}\{\loss(Ax^{(1)},y) + \alpha \loss(x^{(2)},x^{(3)})\},
\end{equation}
where the first term is for data consistency and the second term is to impose equivariance, $\alpha$ is the trade-off parameter to control the strength of equivariance, and $\loss$ is an error function.

\begin{figure*}[!t]
\begin{center}
\includegraphics[width=0.8\linewidth]{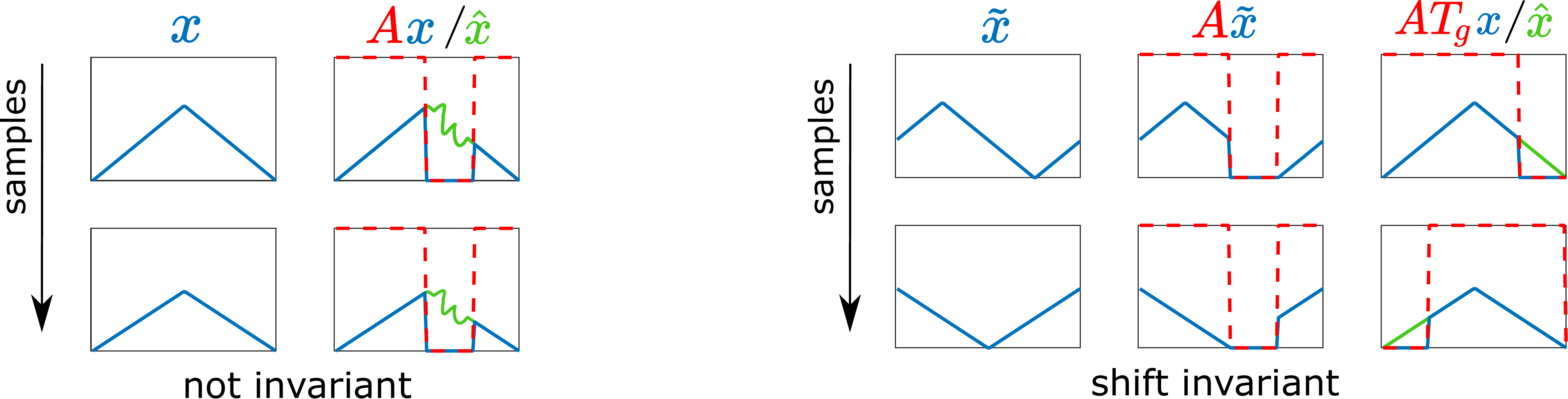}
\end{center}
\captionsetup{font={small}}
\caption{Learning with and without equivariance in a toy 1D signal inpainting task. The signal set consists of different scaling of a triangular signal. On the left, the dataset does not enjoy any invariance, and hence it is not possible to learn the data distribution in the nullspace of $A$. In this case, the network can inpaint the signal in an arbitrary way (in green), while achieving zero data consistency loss. On the right, the dataset is shift invariant. The range of $A$ is shifted via the transformations $T_g$, and the network inpaints the signal correctly.  }
\label{fig:invariance_toy}
\end{figure*}

After training, the learned reconstruction function $f_\theta=G_\theta\circ A^{\dagger}$ can be directly deployed either on the training samples of observations or on new previously unseen observations to recover their respective ground-truth signals. Algorithm \ref{algo:ten} provides the pseudo-code of the \emph{Equivariant Imaging (EI)} where $\loss$ is the mean squared error (MSE).

\paragraph{Adversarial extension}
We can add an additional penalty to enforce the invariant distribution consistency  \eqref{eq:distrib_inv}. 
Inspired by generative adversarial networks~\cite{goodfellow2014generative}, we use a discriminator network $D$ and adopt an adversarial training strategy to further enforce that $x^{(1)}$ and $x^{(2)}$ are identically distributed. The resulting adversarial equivariance learning strategy consists in solving the following optimization:
\begin{equation}\label{eqs:adv_ten_loss}
    \begin{split}
      \min_{G}\max_{D} \mathbb{E}_{y,g}\{\loss(Ax^{(1)},y) &+ \alpha \loss(x^{(2)},x^{(3)})\\&+ \beta \loss_\textrm{adv}(x^{(1)},x^{(2)})\},
    \end{split}
\end{equation}
and $\loss_\textrm{adv}(x^{(1)},x^{(2)})=\mathbb{E}_{x^{(1)}}\{D(x^{(1)})\} + \mathbb{E}_{x^{(2)}}\{1-D(x^{(2)})\}$ \iee~a least square adversarial loss \cite{mao2017least} is adopted and $\beta$ is to control the strength of invariant distribution constraint. Our experimental findings (see in Supplemental material) suggest that the adversarial invariance learning only provides a very slight improvement against the equivariant learning in \eqref{eqs:ten_loss}. Thus in the next sections we mainly focus on equivariant learning.


\section{Theoretical analysis} \label{sec:theory}
We start with some basic definitions. A measurement operator $A\in\eR^{m\times n}$ with $m<n$ has a non-trivial linear nullspace $\nullA \subseteq\mathbb{R}^{n}$ of dimension at least $n-m$, such that $\forall v\in \nullA$ we have $Av=0$.
The complement of $\nullA$ is the range space $\range=\text{range}(A^{\top})$, such that $\range\oplus \nullA=\mathbb{R}^{n}$, which verifies that $\forall v\in\range$ we have $Av\neq 0$.


\vspace{-3pt}
\paragraph{Learning without invariance}
The problem of learning the signal set $\mathcal{X}$ only using compressed samples was first explored in the context of blind compressive sensing~\cite{gleichman2011blind}, for the special case where $\mathcal{X}$ is modelled with a sparse dictionary. The authors in~\cite{gleichman2011blind} showed that learning is impossible in general, becoming only possible when  strong assumptions on the set of plausible dictionaries are imposed.
A similar result can be stated in a more general setting, showing that there are multiple possible reconstruction functions $f$ that satisfy measurement consistency:

\noindent\textbf{Proposition 1} Any reconstruction function $f(y):\mathbb{R}^{m}\mapsto\mathbb{R}^{n}$ of the form
\begin{equation}
\label{eq:unlearnable}
    f(y) = A^{\dagger}y + v(y)
\end{equation}
where $v(y):\mathbb{R}^{m}\mapsto\mathcal{N} $ is any function whose image belongs to the nullspace of $A$ verifies the measurement consistency requirement.

\noindent\textit{Proof:} For $f$ any form \eqref{eq:unlearnable} the measurement consistency can be expressed as $Af(y) = AA^{\dagger}y + Av(y)$ where the first term is simply $y$ as $AA^{\dagger}$ is the identity matrix, and $Av(y)=0$ for any $v(y)$ in the nullspace of $A$. $\square$

For example, the function $v(x)$ can be as simply as $v=0$ and the resulting $f$ will be measurement consistent.
Interestingly, some previous supervised approaches~\cite{schwab2018deepnull,chen2020decomposition} separate the learning of the range and the nullspace components. Proposition 1 shows that without ground truth signals, there is no information to learn the nullspace component.

\paragraph{Learning with invariance}
In the proposed equivariant imaging paradigm, each observation can equally be thought of as a new observation with a new measurement operator $A_g = AT_g$, as
\begin{align}
    y &= A x = AT_g T_g^{\top}x = A_g \tilde{x}
\end{align}
where $\tilde{x}=T_g^{\top}x$ is also a signal in $\mathcal{X}$. Hence, the invariance property allows us to see in the range of the operators $A_g$, or equivalently, \emph{rotate} the range space $\range$ through the action of the group $\group$, \ieB
\begin{equation}
    \mathcal{R}_{A_g} =\text{range}(T_g^{\top}A^{\top})= T_g^{\top}\mathcal{R}_A.
\end{equation}
This idea is illustrated in Figure~\ref{fig:invariance_toy} for inpainting a simple 1D signal model. A necessary condition to recover a unique model $\mathcal{X}$ is that the concatenation of operators $A_g$ spans the full space $\mathbb{R}^{n}$:

\noindent\textbf{Theorem 1}: A necessary condition for recovering the signal model $\mathcal{X}$ from compressed observations is that the matrix
\begin{equation}\label{eq: M}
M = \begin{bmatrix}
AT_1 \\
\vdots \\
AT_{\ntransf}
\end{bmatrix}	\in \mathbb{R}^{\ntransf m \times n}
\end{equation}
is of rank $n$.

\noindent\textit{Proof:} Assume the best case scenario where we have an oracle access to the measurements associated with the different transformations of the same signal\footnote{This is also the case for the simplest signal model where $\mathcal{X}$ is composed of a single atom.} $x$, that is $y_g = AT_g x$ for all $g$. Stacking all the measurements together into $\tilde{y}\in \mathbb{R}^{\ntransf m}$, we observe $\tilde{y}=Mx$ and hence $M$ needs to be of rank $n$ in order to recover $x$. $\square$

This necessary condition provides a lower bound on how big the group $\group$ has to be, \ieB at least satisfy $m\ntransf\geq n$. For example, if the model is invariant to single reflections ($\ntransf=2$), we need at least $m\geq n/2$. Moreover, this condition also tells us that the range space $\mathcal{R}_{A}$ cannot be invariant to $\group$.

\noindent\textbf{Corollary 1}: A necessary condition for recovering the signal model $\mathcal{X}$ from compressed observations is that the range $\mathcal{R}_{A}$ with $m<n$ is not invariant to the action of $\group$, i.e., there is $g\in\group$ such that
\begin{equation}
\mathcal{R}_{A} \neq \mathcal{R}_{AT_g}
\end{equation}
\noindent\textit{Proof:} If $\mathcal{R}_{A} = \mathcal{R}_{AT_g}$ for all $g$ then $\mathcal{N}_{A} = \mathcal{N}_{AT_g}$ for all $g$. From \eqref{eq: M} we have that $M$ shares the same null space and is therefore rank $m<n$. $\square$

Corollary 1 tells us that not any combination of $A$ and $\group$ is useful for learning beyond the range space. For example, shift invariance cannot be used to  learn from Fourier based measurement operators (which is the case in deblurring, super-resolution and magnetic resonance imaging), as $A^{\top}$ would be invariant to the shifts.
It is worth noting that the necessary condition in Theorem 1 will in general not be sufficient. For example, for shift invariant models, a forward matrix composed of a single localized measurement $A=[1,0,\dots,0]^{\top}$ verifies the necessary condition but might not enough to learn a complex model $\mathcal{X}$.




\section{Experiments}
We show experimentally the performance of the proposed method for diverse image reconstruction problems. Due to space limitations, we present a few examples here and include more in the Supplementary Material (SM).

\subsection{Setup and Implementation}
We evaluate the proposed approach on two inverse imaging problems: \emph{sparse-view CT image reconstruction} and \emph{image inpainting}, where the measurement operator $A$ in both tasks are fixed and have non-trivial nullspaces, illustrating the models' ability to learn beyond the range space. We designed our experiments to address the following questions: (i) how well does the equivariant imaging paradigm compare to fully supervised learning? (ii) how does it compare to measurement consistent only learning (\iee~with the equivariance loss term removed)?


\begin{figure*}[t]
\begin{minipage}{1\linewidth}
\centerline{\includegraphics[width=1\textwidth]{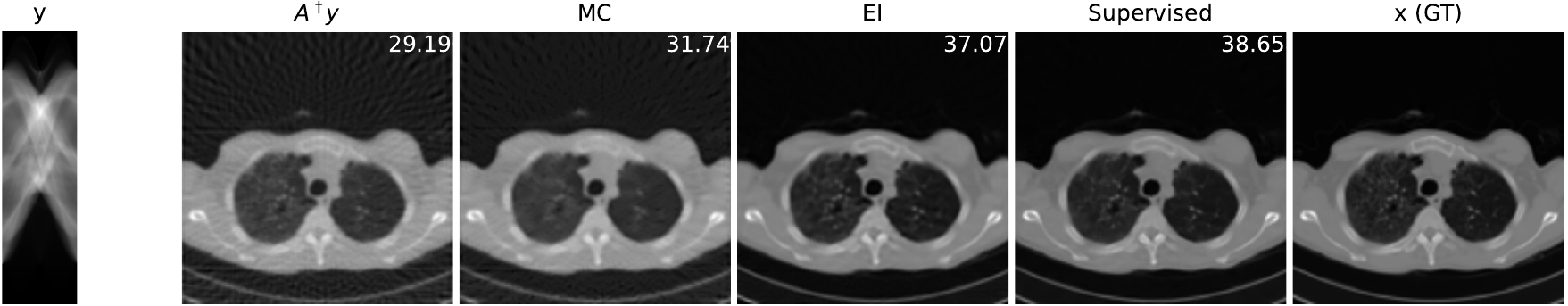}}
\end{minipage}
\begin{minipage}{1\linewidth}
\centerline{\includegraphics[width=1\textwidth]{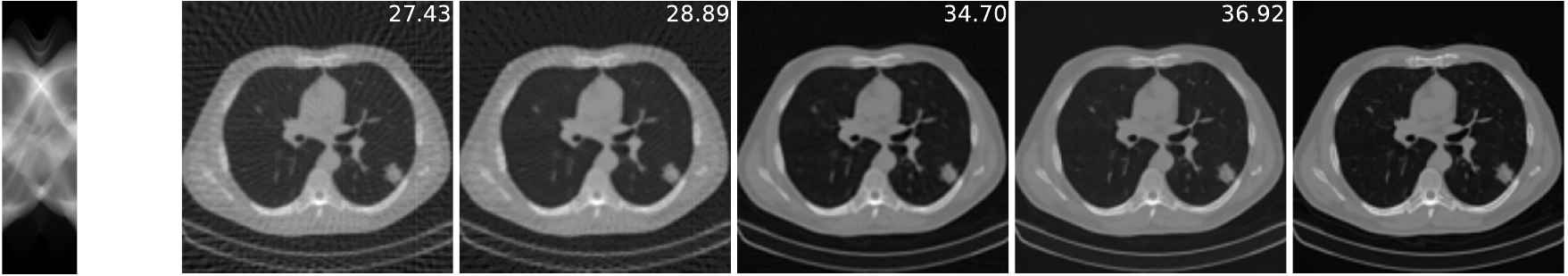}}
\end{minipage}
\begin{minipage}{1\linewidth}
\centerline{\includegraphics[width=1\textwidth]{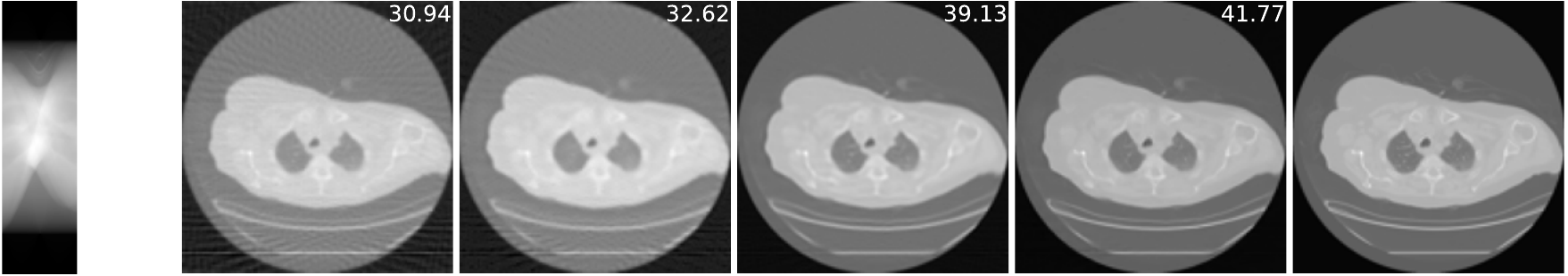}}
\end{minipage}
\begin{minipage}{1\linewidth}
\centerline{\includegraphics[width=1\textwidth]{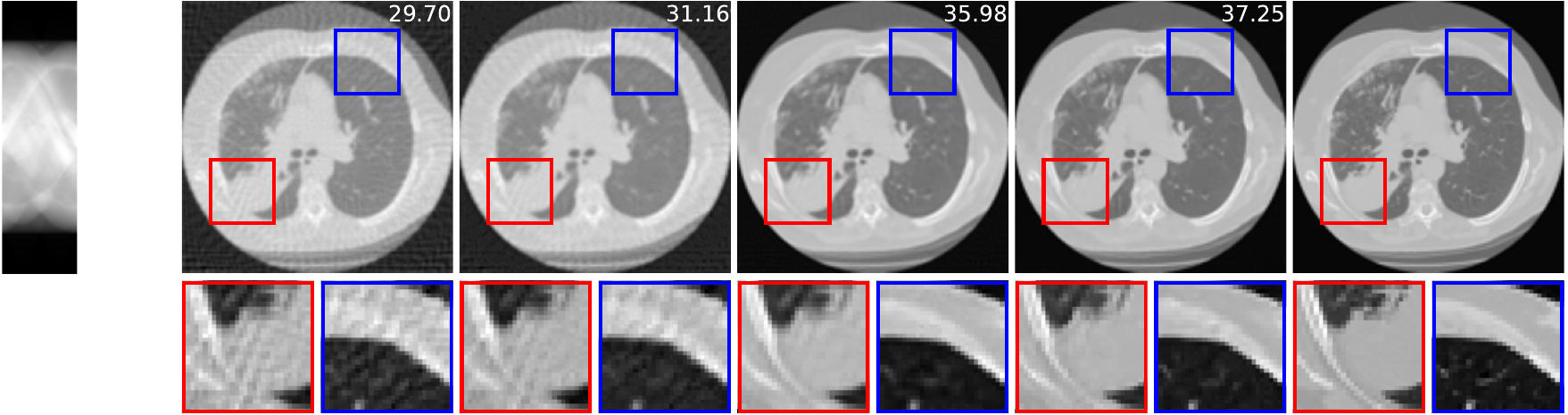}}
\end{minipage}
\caption{Examples of sparse-view CT image reconstruction on the unseen test observations. We train the supervised model (FBPConvNet \cite{jin2017deep}) with observation-groundtruth pairs while train our equivariance learned model with observations alone. We adopt the \emph{random rotation} as the transformation $T$ for our equivariance learning. We obtained results comparable to supervised learning in artifacts-removal. Corresponding PSNR are shown in images.}
\label{fig:ct100}
\end{figure*}

\begin{figure*}[t]
  \centering
    \begin{minipage}[c]{.65\linewidth}
    \centering
    \includegraphics[width=\linewidth]{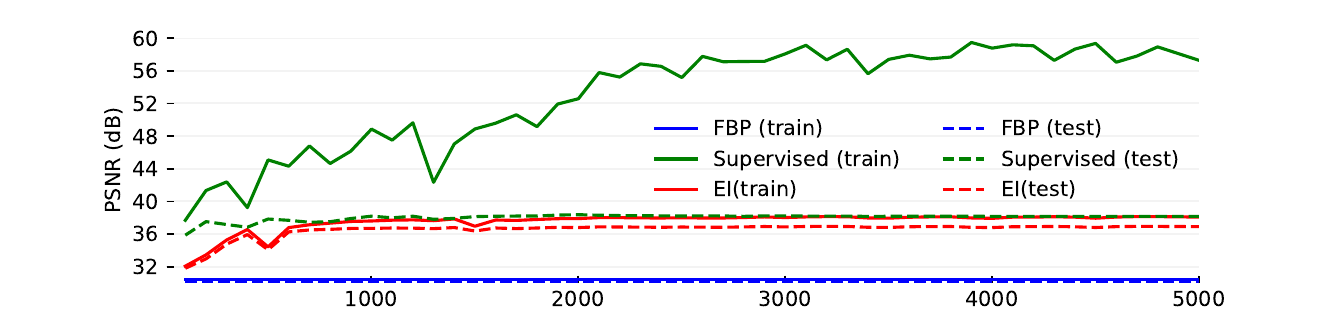}
  \end{minipage} \hspace{0.15pt}
      \begin{minipage}[c]{.32\linewidth}
    \centering
    \captionsetup{font={small}}
\caption{Reconstruction performance (PSNR) as a function of training epoch for supervised trained FBPConvNet and our method (learn without groundtruth) on sparse-view CT observations for training and testing.}
    \label{fig:mse_ct}
  \end{minipage}%
\end{figure*}

Throughout the experiments, we use a U-Net~\cite{ronneberger2015u} to build $G_\theta$ with a residual connection at the output, \iee~$G_\theta = I + G^{\textrm{res}}_\theta$ and $f_\theta(y) = \eeAT y + G^{\textrm{res}}_\theta(\eeAT y)$, to explicitly let the learning target of $G^{\textrm{res}}_\theta$ recover the nullspace component of $x$. We compare our method (EI) with four different learning strategies: measurement-consistency only (MC) with the equivariance term in (\ref{eqs:ten_loss}) removed; the adversarial extension of EI ($\text{EI}_{\text{adv}}$) in \eqref{eqs:adv_ten_loss} using the discriminator network from \cite{wang2018esrgan}; supervised learning (Sup) \cite{jin2017deep} that minimizes $\mathbb{E}_y\{\mathcal{L}(f_\theta(y),x)\}$ using ground truth signal-measurement pairs; and EI regularized supervised learning ($\text{EI}_{\text{sup}}$) with the data consistency term replaced by $\mathcal{L}(f_\theta(y),x)$ in (\ref{eqs:ten_loss}). For a fair comparison with EI, no data augmentation of ground truth signals are conducted for both supervised learning methods, Sup and $\text{EI}_{\text{sup}}$. We use the residual U-Net architecture  for all the counterpart learning methods to ensure all methods have the same inductive bias from the neural network architecture. Note that while there are many options to determine the optimal network architecture such as exploring different convolutions~\cite{ren2015shepard,yu2019free,liu2018image,yu2018generative} or different depths~\cite{ulyanov2018deep}, these aspects are somewhat orthogonal to the \emph{learning beyond the range space} question.

We demonstrate that the equivariant imaging approach is straightforward and can be easily extended to existing deep models without modifying the architectures. All methods are implemented in PyTorch and optimized by Adam~\cite{kingma2014adam}. We tuned the $\alpha$ for specific inverse problems (see SM for training details).



\begin{figure*}[t]
\begin{minipage}{1\linewidth}
\centerline{\includegraphics[width=1\textwidth]{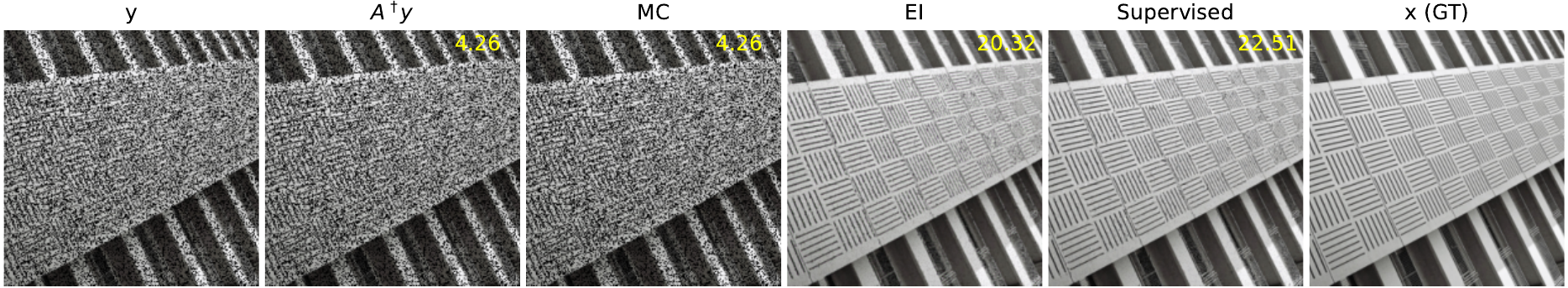}}
\end{minipage}
\begin{minipage}{1\linewidth}
\centerline{\includegraphics[width=1\textwidth]{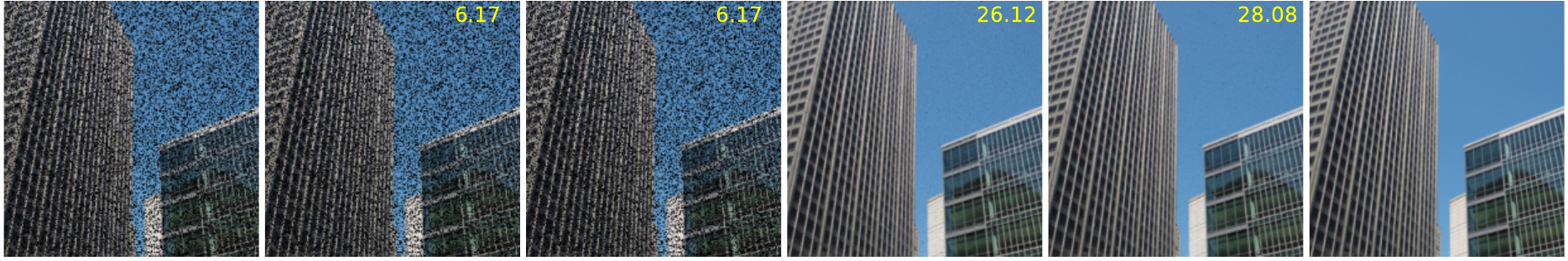}}
\end{minipage}
\begin{minipage}{1\linewidth}
\centerline{\includegraphics[width=1\textwidth]{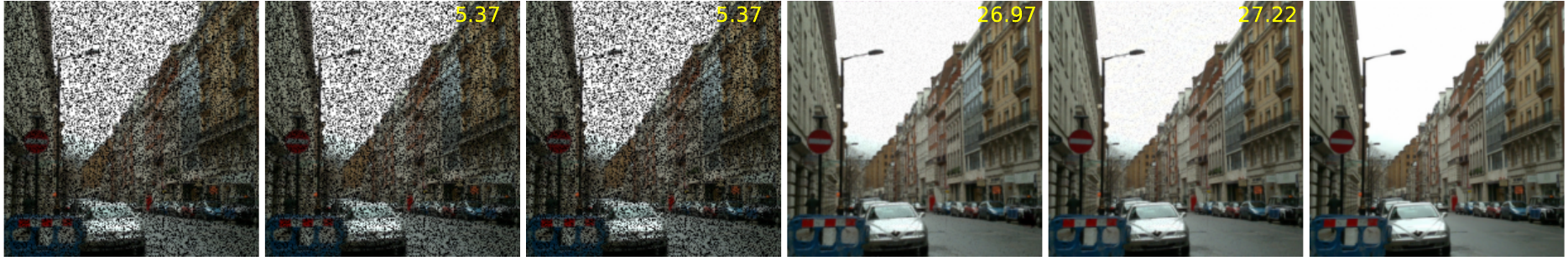}}
\end{minipage}
\begin{minipage}{1\linewidth}
\centerline{\includegraphics[width=1\textwidth]{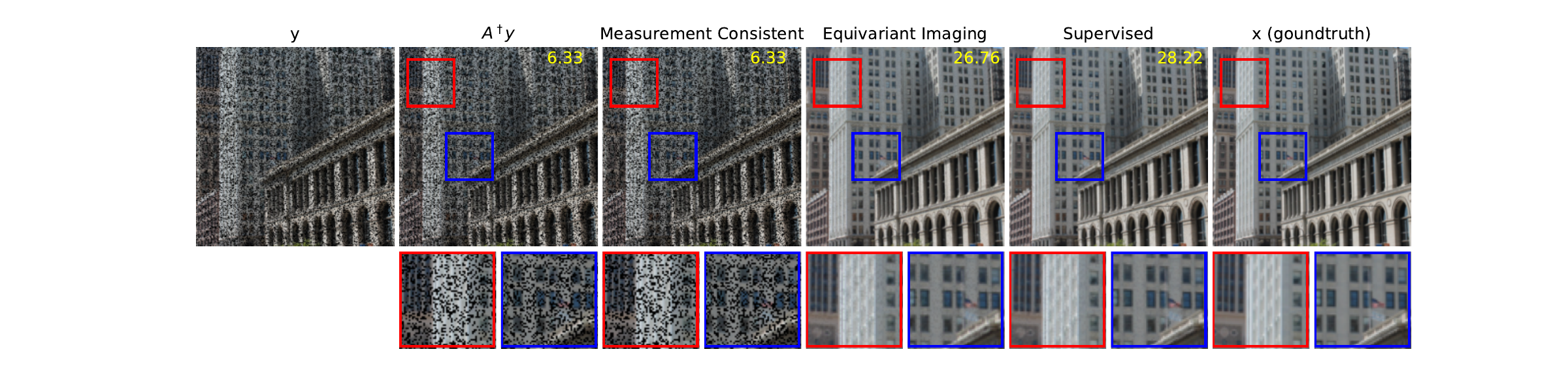}}
\end{minipage}
\caption{Examples of pixelwise image inpainting on the unseen test observations. We train the supervised model \cite{jin2017deep} with observation-groundtruth pairs while train our equivariance learned model with observations alone. We adopt the \emph{random shift} as the transformation $T$ for our equivariance learning. We obtained results comparable to supervised learning in recovering missing pixels. Corresponding PSNR are shown in images.}
\vspace{-10pt}
\label{fig:urban100}
\end{figure*}

\subsection{Sparse-view CT}
The imaging physics model of X-ray computed tomography (CT) is the discrete \texttt{radon} transform. The physics model $A$ is the \texttt{radon} transformation where 50 views (angles) are uniformly subsampled to generate the sparse-view sinograms (observations) $y$. The Filtered back projection (FBP) function, \iee~\texttt{iradon}, is used to approximate $\eeAT$. In this task, we exploit the invariance of the CT images to rotations\footnote{It is worth noting that shift invariance is not useful for the CT case, as the forward operator is shift invariant itself (see Corollary 1).}, and $\group$ is the group of rotations by $1$ degree ($\ntransf$=360).
We use the CT100 dataset \cite{clark2013cancer}, a public real CT clinic dataset which comprises 100 real in-vivo CT images collected from the cancer imaging archive\footnote{\url{https://wiki.cancerimagingarchive.net/display/Public/TCGA-LUAD}} which consist of the middle slice of CT images taken from 69 different patients. The CT images are resized to $128\times 128$ pixels and we then apply the \texttt{radon} function on them to generate the $50$-views sinograms. We used the first 90 sinograms for training while the remaining 10 sinograms for testing. Note in this task, the supervised trained residual U-Net is just the FBPConvNet proposed in~\cite{jin2017deep} which has been demonstrated to be very effective in supervised learning for sparse-view CT image reconstruction.  We train our model with equivariance strength $\alpha=10^2$ (see SM for more results and the equivariance strength effect). using the sinograms $y$ alone while the FBPConvNet is trained with the ground truth pairs $(x,y)$.

A qualitative comparison is presented in Figure~\ref{fig:ct100}. The sparse-view FBP contains the line artifacts. Both the FBPConvNet and our methods significantly reduce these artifacts, giving visually indistinguishable results. Figure~\ref{fig:mse_ct} shows the value of PSNR of reconstruction on the training measurements and test measurements and we have the following observations: (i) We would naturally expect the network trained with ground truth data to perform the best. However, we note that the equivariant test error is almost as good despite having no access to ground truth images and only learning on the sparse sinogram data. Furthermore the EI solution is about 7 dB better than the FBP, clearly demonstrating the correct learning of the null space component of the image. (ii) We note that there is a significant gap between training and test error for the  FBPConvNet, suggesting that the network may be overfitting. We do not observe this in the EI learning. This can be explained by the fact that the EI constrains the network to a much small class of functions (those that are equivariant on the data) and thus can be expected to have better generalization properties.  

We also compared the EI with its adversarial extension in (\ref{eqs:adv_ten_loss}) and the supervised learning regularized by equivariance objective. The quantitative results are given in table \ref{table:compare} below. First, MC learning obtains a small improvement in performance over FBP which may be attributable to the fact that FBP is only an approximation to $\eeAT$. Alternatively it may be due to the inductive bias of the neural network architecture~\cite{tachella2020cnn}.
Second, the adversarial extension provides a slight improvement to EI and similarly the EI regularization helps the vanilla supervised learning obtain a further 0.6 dB improvement. These results suggest that it is indeed possible to learn to reconstruct challenging inverse problems with only access to measurement data.


\begin{table}[t]
\begin{center}
\fontsize{8}{12}\selectfont
\begin{tabular}{c|cccccc}
& FBP& MC & EI& $\text{EI}_{\text{adv}}$& Sup & $\text{EI}_{\text{sup}}$\\
\hline
50-views CT & 30.24 &31.01& 36.94& 36.96& 38.17& 38.79
\end{tabular}
\begin{tabular}{c|cccccc}
& $\eeAT y$& MC & EI& $\text{EI}_{\text{adv}}$& Sup & $\text{EI}_{\text{sup}}$\\
\hline
Inpainting & 5.84& 5.84 &25.14 &23.26 & 26.51& 26.75\\
\end{tabular}
\end{center}
\caption{Reconstruction performance (PSNR) of 50-views CT reconstruction  and image inpainting  for different methods on the CT100  and Urban100 test measurements, respectively.}\label{table:compare}
\end{table}




\subsection{Image inpainting}
As a proof-of-concept of the generality of the method, we also applied our method on an image inpainting task with a fixed set of deleted pixels. This is relevant for example to the problem of reconstructing images from cameras with hot or dead pixels.

In the image inpainting task, the corrupted measurement is given by $y=b\odot x$ where $b$ is a binary mask, $\odot$ is the Hadamard product, and the associated operator $A=\text{diag}(b)$ and $A=\eeAT$. Here, we consider \emph{pixelwise inpainting} where we \emph{randomly} drop $30\%$ of pixel measurements. We train our model by applying \emph{random shift} transformations. We evaluate the reconstruction performance of our approach and other learning methods using the Urban100 \cite{Huang-CVPR-2015} natural image dataset. For each image, we cropped a 512x512 pixel area at the center and then resized it to 256x256 for the ground truth image. The first 90 measurements are for training while the last 10 measurements are for testing.

The reconstruction comparisons are presented in Figure~\ref{fig:urban100} and Table~\ref{table:compare}. We have the following observations: First, the MC reconstruction is exactly equal to $\eeAT y$ as the exact pseudo inverse is used, the reconstruction quality of MC is very poor as it completely failed to learn the nullspace at all. Second, the EI reconstruction is about 20 dB better than $\eeAT y$ and MC reconstruction, the missing pixels are recovered well, again demonstrating the correct learning of the null space component of the image (the adversarial EI was not competitive in this application). Finally, there is only a 1.37 dB gap between the reconstruction of EI and the fully supervised model. As with the CT imaging, we again find the generalization error of EI is also much smaller than for the supervised model (see SM).

\section{Discussion}
The equivariant imaging framework presented here is conceptually different from recent ideas on invariant networks~\cite{lenc2015understanding,schmidt2012learning,foster2020improving} where the goal is to train an invariant neural network for classification problems, which generally performs better than a non-invariant one~\cite{sokolic2017generalization}. In contrast, the equivariant imaging goal is to make the composition $f_\theta\circ A$ equivariant but not necessarily $f_\theta$, promoting invariance across the complete imaging system.
Moreover, our framework also differs from standard data augmentation techniques, as no augmentation can be done directly on the compressed samples $y$. The proposed method overcomes the fundamental limitation of only having range space information, effectively solving challenging inverse problems without the need of ground truth training signals. 
As shown in the experiments, our equivariant constraint can also be applied in the fully supervised setting to improve the performance of the networks.



The equivariant imaging framework admits many straightforward extensions.
 For example, while we have shown how to use shift-invariance and rotation-invariance to solve the inpainting task and CT reconstruction. We believe that there are many other imaging tasks that could benefit from equivariant imaging. It would also be very interesting to investigate whether the benefits seen here can be extended to nonlinear imaging problems.

\emph{Mixed types of group transformations} can also be applied at the training time and may help improve convergence time and performance. However, as we have shown, the strength of different transformations will depend on the nature of the signal model and the physics operator.

We have also found that equivariant imaging can be used to improve the performance for \emph{single image reconstruction} and have reported some preliminary results in the Supplementary material. However, as single image reconstruction itself relies heavily on the strong inductive bias of the network~\cite{tachella2020cnn} the role is EI in this scenario is less clear.


\section{Conclusions}
We have introduced a novel self-supervised learning strategy that can learn to solve an ill-posed inverse problem from only the observed measurements, without having any knowledge of the underlying signal distribution, other than assuming that it is invariant to the action of a group of transformations. This relates to an important question on the use of deep learning in scientific imaging~\cite{Belthangady2019}: can networks learn to image structures and patterns for which no ground truth images yet exist? We believe that the EI framework suggests that with the addition of the basic physical principle of invariance, such data-driven discovery is indeed possible.

\section*{Acknowledgments}
This work is supported by the ERC C-SENSE project (ERCADG-2015-694888).

\bibliographystyle{IEEEtran}
\bibliography{egbib}

\newpage
\appendix
\section{Training Details} \label{sec:details}
\begin{figure}[h]
\begin{center}
\includegraphics[width=1\linewidth]{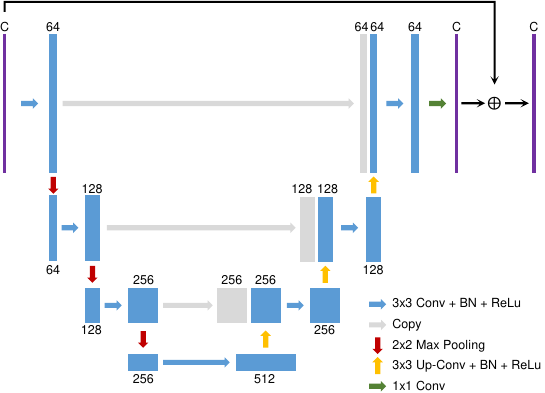}
\end{center}
\caption{The residual U-Net \cite{ronneberger2015u} used in the paper. The number of input and output channels is denoted as $C$, with $C=1$ in the CT task and $C=3$ in the inpainting task.}
\label{fig:unet}
\end{figure}

We first provide the details of the network architectures and hyperparameters of Figs. 4-6 and Table 1 of the main paper. We implemented the algorithms and operators (\egg~\texttt{radon} and \texttt{iradon}) in Python with PyTorch 1.6 and trained the models on NVIDIA 1080ti and 2080ti GPUs.
Figure \ref{fig:unet} illustrates the architecture of the residual U-Net used  \cite{ronneberger2015u} in our paper.

For the sparse-view CT task, we used the Adam optimizer with a batch size of $2$ and an initial learning rate of $0.0005$. The weight decay is $10^{-8}$. The distribution strength $\beta$ is $10^{-8}$ for $\text{EI}_{adv}$. We trained the networks for $5000$ epochs, keeping the learning rate constant for the first 2000 epochs and then shrinking it by a factor of $0.1$ every 1000 epochs. More reconstruction examples are presented in Figure \ref{fig:ct100_SM}. 

\begin{figure}[h]
\begin{center}
\includegraphics[width=1\linewidth]{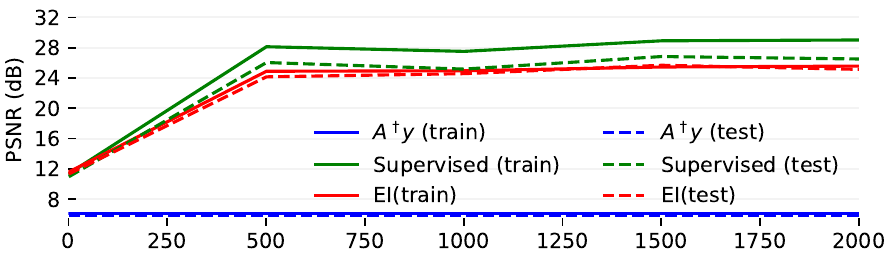}
\end{center}
\caption{Reconstruction performance (PSNR) as a function of training epoch for the supervised model \cite{jin2017deep} and our method (no ground truth) on inpainting task measurements for training and testing.}
\label{fig:psnr_ipt}
\end{figure}

\begin{figure*}[t]
\begin{minipage}{1\linewidth}
\centerline{\includegraphics[width=1\textwidth]{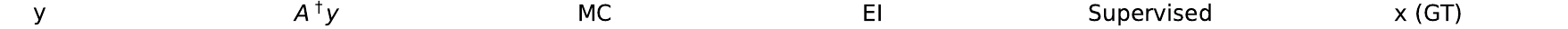}}
\end{minipage}
\begin{minipage}{1\linewidth}
\centerline{\includegraphics[width=1\textwidth]{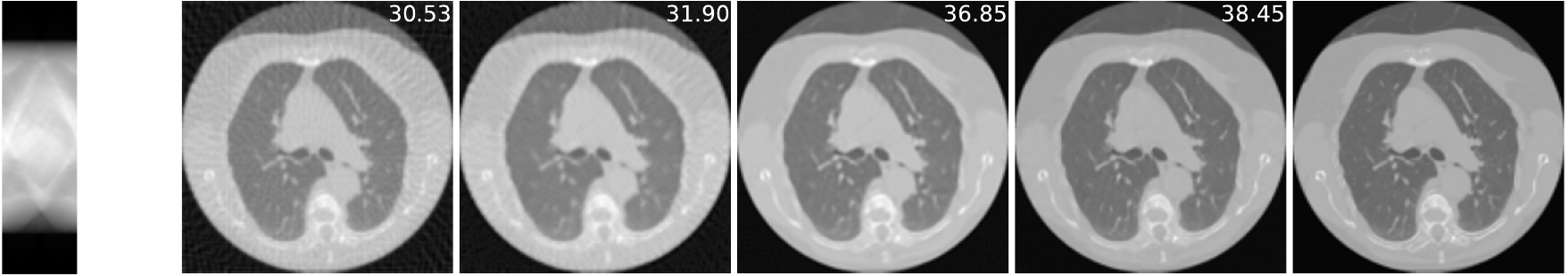}}
\end{minipage}
\begin{minipage}{1\linewidth}
\centerline{\includegraphics[width=1\textwidth]{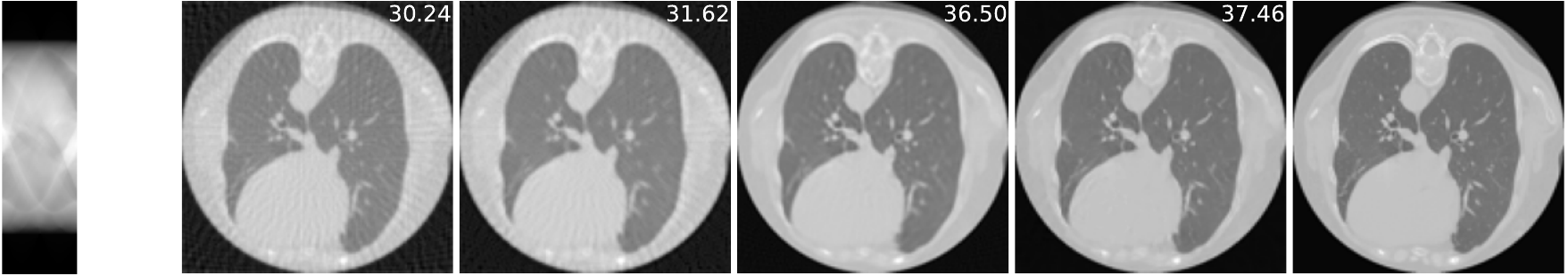}}
\end{minipage}
\begin{minipage}{1\linewidth}
\centerline{\includegraphics[width=1\textwidth]{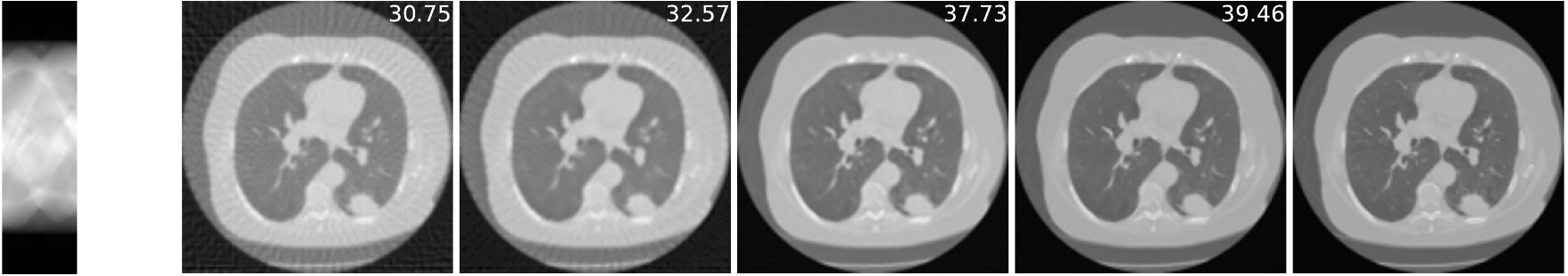}}
\end{minipage}
\caption{More examples of sparse-view CT image reconstruction on the unseen test measurements. We train the supervised model (FBPConvNet \cite{jin2017deep}) with measurement/ground truth pairs while we train the equivariance learned model with measurements alone. We adopt \emph{random rotations} as the transformation $T$ for our equivariance learning. We obtained results comparable to supervised learning in artifacts-removal. Corresponding PSNR are shown in images.}
\label{fig:ct100_SM}
\end{figure*}

\begin{figure*}[t]
\begin{minipage}{1\linewidth}
\centerline{\includegraphics[width=1\textwidth]{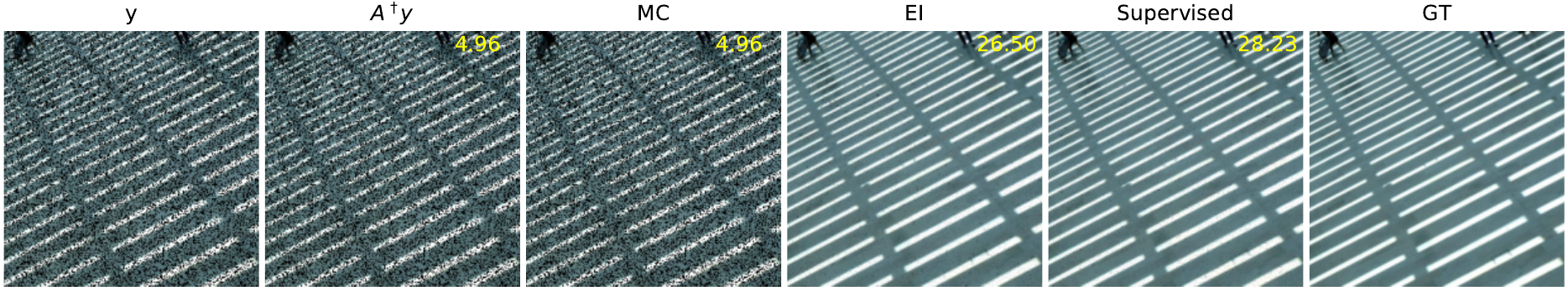}}
\end{minipage}
\begin{minipage}{1\linewidth}
\centerline{\includegraphics[width=1\textwidth]{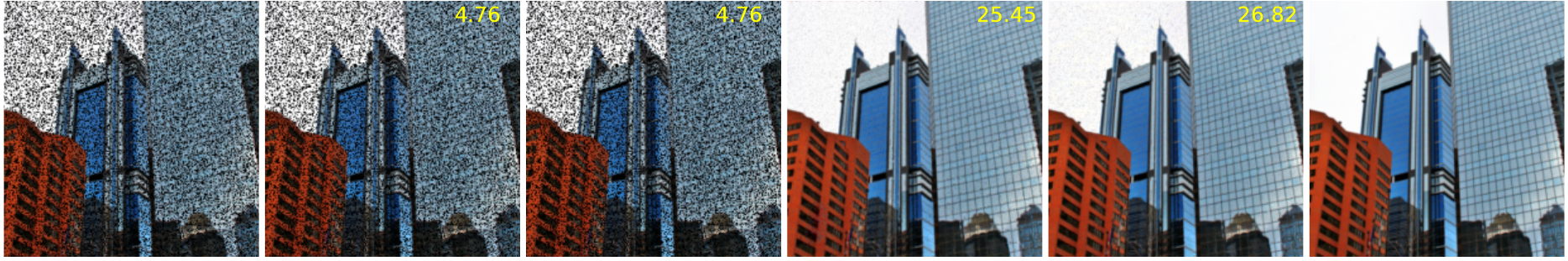}}
\end{minipage}
\begin{minipage}{1\linewidth}
\centerline{\includegraphics[width=1\textwidth]{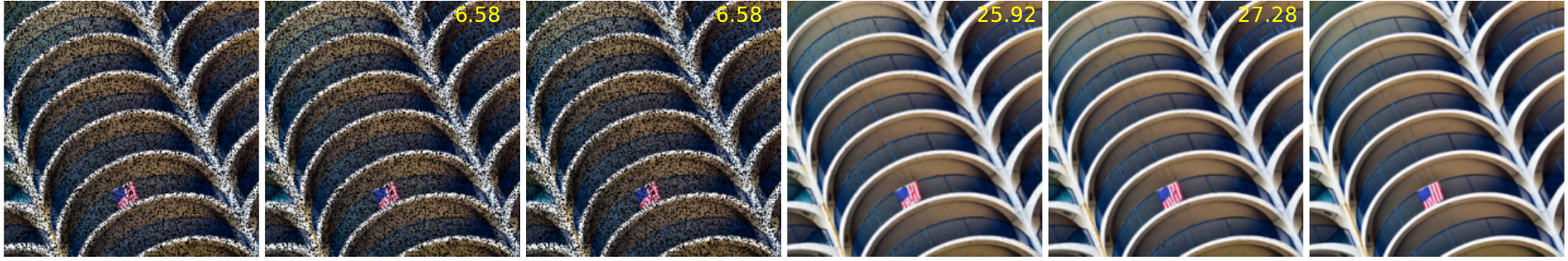}}
\end{minipage}
\caption{More examples of image inpainting reconstruction on the unseen test measurements. We train the supervised model~\cite{jin2017deep} with measurement/ground truth pairs while we train the equivariance learned model with measurements alone. We adopt \emph{random shifts} as the transformation $T$ for our equivariance learning. We obtained results comparable to supervised learning in recovering missing pixels. Corresponding PSNR are shown in images.}
\label{fig:inpainting_SM}
\end{figure*}

For the inpainting task, we also used Adam but with a batch size of 1 and an initial learning rate of $0.001$. The weight decay is $10^{-8}$. The distribution strength $\beta$ is $10^{-8}$ for $\text{EI}_{adv}$. We trained the networks for $2000$ epochs,  shrinking the learning rate by a factor of $0.1$ every 500 epochs.  Figure \ref{fig:psnr_ipt} shows  the peak signal-to-noise ratio (PSNR)  of the reconstructions  on  the  training and test measurements. Again, the generalization error of EI is smaller than for the supervised model. More reconstruction examples are presented in Figure \ref{fig:inpainting_SM}.

\section{More results}
\paragraph{Effect of the equivariance hyperparameter $\alpha$ } Table \ref{table:compare_SM} shows EI reconstruction performance (PSNR) with different equivariance strength  values ($\alpha$ in Eqn. (6) of the main paper). It performs reasonably well when $\alpha=100$ for the CT task and $\alpha=1$ for the inpainting task. When $\alpha$ is too small, the performance drops considerably; at the extreme of no equivariance $(\alpha=0)$, the model fails to learn. These results support our motivation of equivariant imaging.

\begin{table}[h]
\begin{center}
\fontsize{8}{12}\selectfont
\begin{tabular}{c|ccccc}
$\alpha$ &0  & 1& 10 & 100 & 1000 \\
\hline
50-views CT  &31.01 & 36.78 & 36.88 & 36.94& 33.31
\end{tabular}
\begin{tabular}{c|cccc}
$\alpha$ &0  & 0.1& 1 & 10\\
\hline
Inpainting & 5.84  & 23.42 &25.14 &  22.96\\
\end{tabular}
\end{center}
\caption{Effect of the equivariance hyperparameter $\alpha$  on the reconstruction performance (PSNR) in the 50-views CT reconstruction (CT100 dataset) and image inpainting (Urban100 dataset) tasks.}\label{table:compare_SM}
\end{table}

\paragraph{Effect of the networks' inductive bias}
In the deep image prior (DIP) paper, the authors showed that some specific convolutional networks can be  trained to fit a single image by only enforcing measurement consistency~\cite{ulyanov2018deep}. The DIP approach relies heavily on the choice of the network architecture (generally an autoencoder), and does not work with  various popular architectures (\egB those with skip-connections). Moreover, this approach is constrained to a single image and cannot incorporate additional training data.

\begin{figure*}[t]
\begin{minipage}{1\linewidth}
\centerline{\includegraphics[width=1\textwidth]{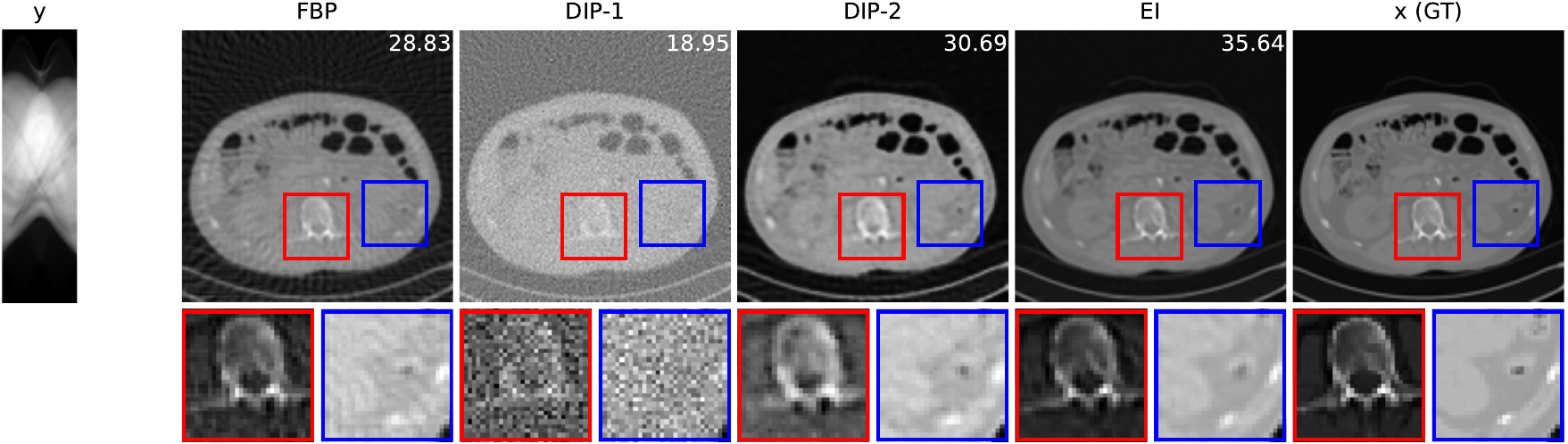}}
\end{minipage}
\caption{Comparison between EI and DIP on 50-views CT reconstruction. We denote DIP-1 and DIP-2 as the DIP learned models trained with residual U-Net (same as EI) and Encoder-Decoder (the best architecture for DIP as suggested in \cite{ulyanov2018deep}), respectively. We trained EI on a measurement set and direct apply the trained model on the given new measurement here. Both DIP methods are trained using the given measurement here. Corresponding PSNR are shown in images.}
\label{fig:ct100_DIP_EI}
\end{figure*}

In contrast, we show that our method can learn beyond the range space without heavily relying on the inductive bias of an specific autoencoder architecture. Moreover, we show that EI outperforms the best DIP architecture as it leverages the full compressed training dataset.
We compare our method with the DIP on the 50-views CT image reconstruction task. 
For our method, we use the same residual U-Net as in the other experiments. We build the DIP using two architectures: the same residual U-Net used in EI (which we denote DIP-1) and the best autoencoder network suggested in \cite{ulyanov2018deep} (which we denote DIP-2).
Following \cite{ulyanov2018deep}, we input iid Gaussian noise to both DIP-1 (1 channel) and DIP-2 (32 channels). Our model is trained using the hyperparameters for sparse-view CT (see Section \ref{sec:details}). We train DIP-1 and DIP-2 using 5000 training iterations and a learning rate of $0.001$. As shown in Figure \ref{fig:ct100_DIP_EI}, our method outperforms the DIP methods. DIP-2 performs significantly better than DIP-1 due to the inductive bias of that autoencoder architecture. In contrast, our method works very well even with the residual U-Net. Moreover, our model also outperforms DIP-2 by 5 dB.

\begin{figure*}[t]
\begin{minipage}{1\linewidth}
\centerline{\includegraphics[width=1\textwidth]{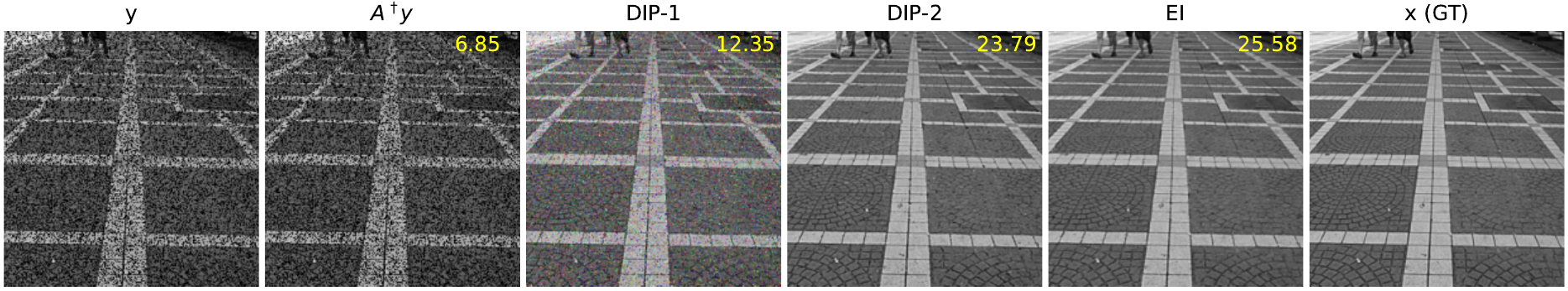}}
\end{minipage}
\begin{minipage}{1\linewidth}
\centerline{\includegraphics[width=1\textwidth]{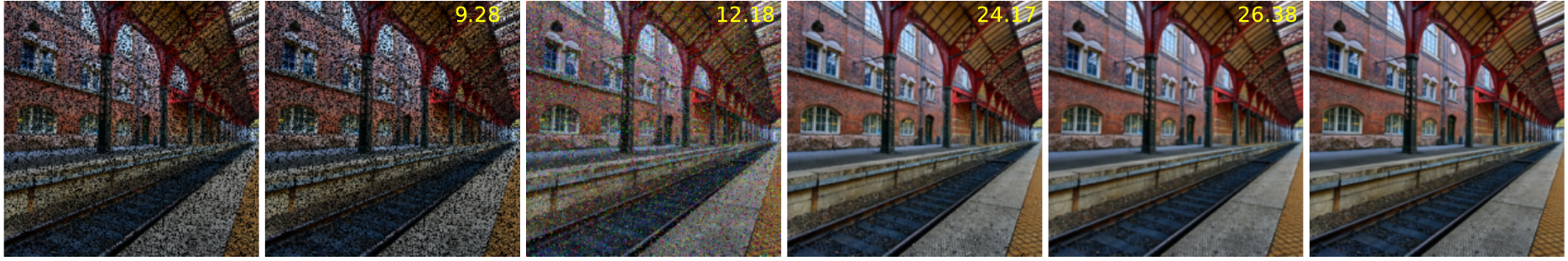}}
\end{minipage}
\caption{Comparison between EI and DIP for single image reconstruction on the inpainting task. We denote DIP-1 and DIP-2 as the DIP learned models trained with residual U-Net (same as EI) and Encoder-Decoder (the best architecture for DIP as suggested in \cite{ulyanov2018deep}), respectively. All the models are trained with the given single compressed measurement data $y$. Corresponding PSNR are shown in images.}
\label{fig:inpainting_vs_dip_SM}
\end{figure*}

\paragraph{Equivariant imaging using a single training image}
We are interested in whether the proposed method works for single image reconstruction, \iee~reconstructing  a single compressed measurement. Here we provide some preliminary results. As an example, we compared our method with the DIP on the inpainting task for single image reconstruction. We trained all 3 models (EI, DIP-1, DIP-2) using 5000 training iterations and a learning rate of $0.001$ on a single measurement input. The results are presented in Figure \ref{fig:inpainting_vs_dip_SM}. We observe that our method works very well for this single image reconstruction task and outperforms both DIP-1 and DIP-2. In addition, DIP-1 performs worse than DIP-2 due to the residual architecture with skip-connections. Again, our model is not so dependent on the inductive bias of network and works well when using the residual connections.
We note that although our method is able to learn with a single measurement, the role of equivariance in this scenario needs to be explored more, and we leave this for future work.

\end{document}